\documentclass[10pt,twocolumn,letterpaper]{article}

\usepackage[pagenumbers]{cvpr} 

\usepackage{graphicx}
\usepackage{amsmath}
\usepackage{amssymb}
\usepackage{booktabs}

\usepackage[pagebackref,breaklinks,colorlinks]{hyperref}
\usepackage{acronym}
\usepackage{subcaption}
\usepackage{array,multirow}

\usepackage[capitalize]{cleveref}
\crefname{section}{Sec.}{Secs.}
\Crefname{section}{Section}{Sections}
\Crefname{table}{Table}{Tables}
\crefname{table}{Tab.}{Tabs.}

\newcommand\blfootnote[1]{%
  \begingroup
  \renewcommand\thefootnote{}\footnote{#1}%
  \addtocounter{footnote}{-1}%
  \endgroup
}

\def\cvprPaperID{5990} 
\def\confName{CVPR}
\def\confYear{2023}

\usepackage[dvipsnames]{xcolor}

\newcommand{\adj}[1]{\textcolor{MidnightBlue}{#1}}
\newcommand{\noun}[1]{\textcolor{OliveGreen}{#1}}
\newcommand{\clause}[1]{\textcolor{Brown}{#1}}

\acrodef{cv}[CV]{Computer Vision}
\acrodef{nlp}[NLP]{Natural Language Processing}
\acrodef{vl}[V\&L]{Vision-Language}
\acrodef{t2i}[T2I]{Text-to-Image}
\acrodef{t2v}[T2V]{Text-to-video}
\acrodef{t2s}[T2S]{Text-to-Simulation}
\acrodef{t2v/s}[T2V/S]{Text-to-Video/Simulation}
\acrodef{t2a}[T2A]{Text-to-Animation}
\acrodef{t2-3d}[T2-3D]{Text-to-3D}
\acrodef{mpm}[MPM]{Material Point Method}
\acrodef{ipc}[IPC]{Incremental Potential Contact}
\acrodef{nerf}[NeRF]{Neural Radiance Fields}
\acrodef{llm}[LLM]{Large-scale Language Model}
\acrodef{gan}[GAN]{Generative Adversarial Network}
\acrodef{ddm}[DDM]{Denoising Diffusion Models}

\usepackage{xspace}
\makeatletter
\DeclareRobustCommand\onedot{\futurelet\@let@token\@onedot}
\def\@onedot{\ifx\@let@token.\else.\null\fi\xspace}
 
\def\ie{\emph{i.e}\onedot} 
 
\def\etc{\emph{etc}\onedot}

\usepackage{hyperref}
\begin{document}

\title{TPA-Net: Generate A Dataset for \underline{T}ext to \underline{P}hysics-based \underline{A}nimation$^{\heartsuit}$}
\author{Yuxing Qiu$^{1}$ \quad Feng Gao$^{1}$ \quad Minchen Li$^{1}$ \quad Govind Thattai \quad Yin Yang$^{2}$ \quad Chenfanfu Jiang$^{1}$ \\
$^1$ University of California, Los Angeles \qquad $^2$ University of Utah\\
{\tt\small \{yxqiu,f.gao\}@ucla.edu, gowin.thattai@gmail.com, yin.yang@utah.edu,}\\ 
{\tt\small \{minchen,cffjiang\}@math.ucla.edu}
}

\maketitle

\blfootnote{$\heartsuit$ Work in Progress.}



\begin{abstract}
    Recent breakthroughs in \acf{vl} joint research have achieved remarkable results in various text-driven tasks. High-quality \acf{t2v}, a task that has been long considered mission-impossible, was proven feasible with reasonably good results in latest works. However, the resulting videos often have undesired artifacts largely because the system is purely data-driven and agnostic to the physical laws. To tackle this issue and further push \ac{t2v} towards high-level physical realism, we present an autonomous data generation technique and a dataset, which intend to narrow the gap with a large number of multi-modal, 3D \acf{t2v/s} data. In the dataset, we provide high-resolution 3D physical simulations for both solids and fluids, along with textual descriptions of the physical phenomena. We take advantage of state-of-the-art physical simulation methods (i) \acf{ipc} and (ii) \acf{mpm} to simulate diverse scenarios, including elastic deformations, material fractures, collisions, turbulence, \etc. Additionally, high-quality, multi-view rendering videos are supplied for the benefit of \ac{t2v}, \acf{nerf}, and other communities. This work is the first step towards fully automated \acf{t2v/s}. Live examples and subsequent work are at \url{https://sites.google.com/view/tpa-net}.
\end{abstract}

\begin{figure*}
    \centering
    \includegraphics[width=\linewidth]{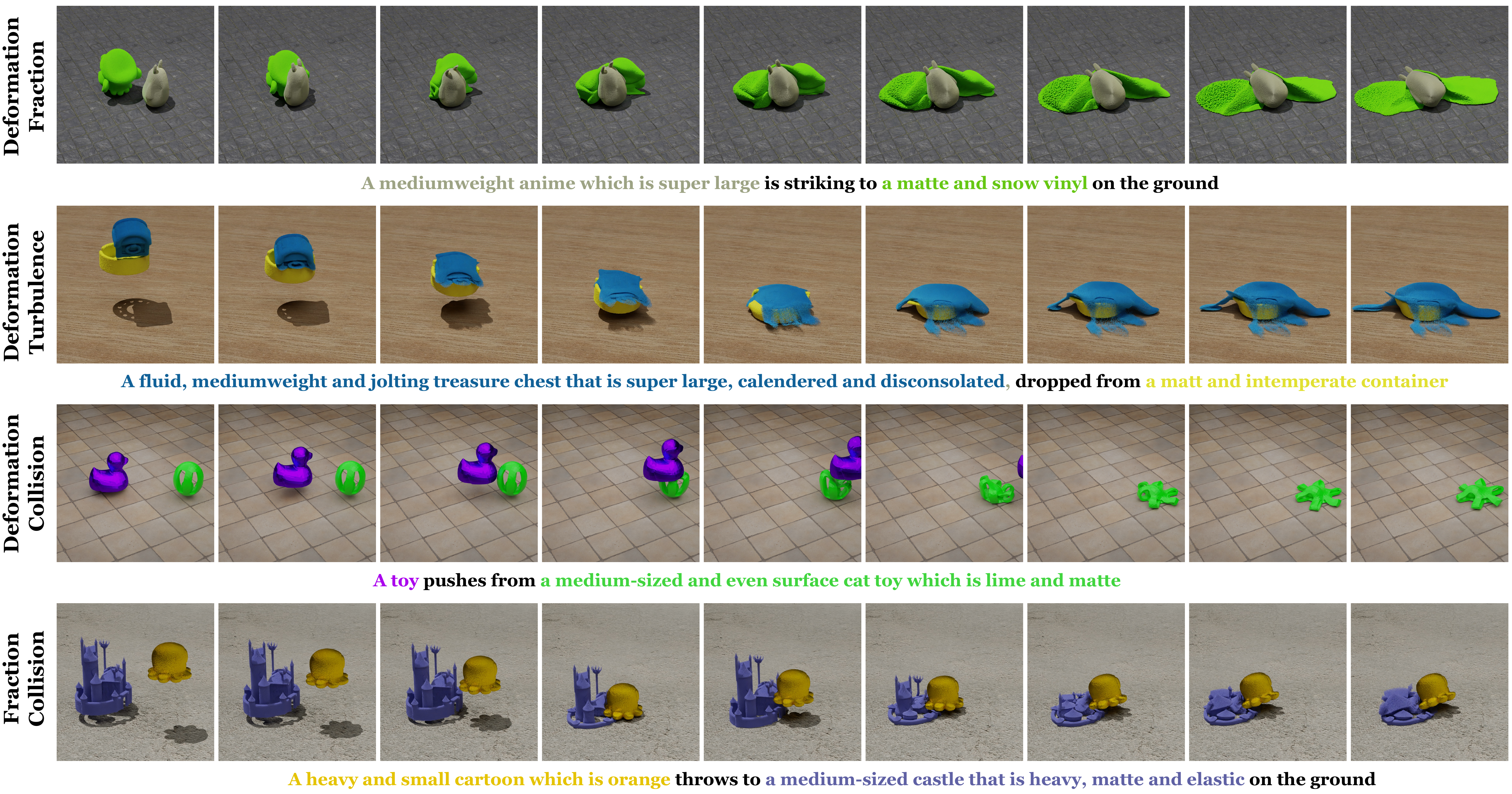}
    \caption{Examples of animation in TPA-Net datase. The animations cover a wide range of physical phenomenons, such as \textbf{deformation}, \textbf{fraction}, \textbf{collision} and \textbf{turbulence} \etc. Each physically realistic animation is described by human readable text.}
    \label{fig:teaser}
\end{figure*}

\section{Introduction}
\label{sec:intro}
In the past years, we have witnessed the blooming of the \acf{vl} community in solving diverse daily-life tasks \cite{lu2019vilbert,li2019visualbert,chen2020uniter,ramesh2021zero,radford2021learning,zhang2021vinvl,alayrac2022flamingo,Gao_2022_CVPR,kamath2021mdetr,ramesh2022hierarchical}. 
Particularly, multi-modal \ac{vl} models have achieved remarkable performances on various conventional \ac{vl} tasks thanks to the availability of an enormous amount of \ac{vl} data and rapidly developing \acf{llm} \cite{vaswani2017attention,devlin2018bert,radford2019language,lewis2019bart,brown2020language}.
With those handy tools, researchers took on more generative problems, such as \acf{t2i}, and came up with impressive solutions \cite{ramesh2021zero,ramesh2022hierarchical}. 
Here, three key facts jointly contribute to these \acf{t2i} successes: 
(i) self-supervised learning techniques and self-attention deep learning architectures are fully explored \cite{vaswani2017attention,devlin2018bert,radford2019language,lewis2019bart,brown2020language};
(ii) \acf{vl} generative models and learning paradigms are well studied \cite{goodfellow2020generative,zhu2017unpaired,ho2020denoising,rombach2022high};
(iii) a large volume of image-text pairs are available on the internet \cite{lin2014microsoft,ordonez2011im2text,plummer2015flickr30k,sharma2018conceptual,changpinyo2021conceptual,srinivasan2021wit} to enable (i) and (ii) to capture the correlations between vision features and language representations. 

However, \acf{t2v} research is more challenging.
First, due to the lack of data, it is much harder to train \acf{t2v} models from scratch \cite{singer2022make}, even though both (i) and (ii) could be applied to \ac{t2v}. 
With no efficient data generation and labeling methods, researchers have to explore other alternatives for \ac{t2v} solutions (using, for example, pre-trained \ac{t2i} models, diffusion-based models, and image priors \cite{singer2022make,ho2022imagen}). 
Second, unlike \acf{t2i}, we cannot make the assumption in \acf{t2v} that the descriptive text contains equal amounts of information as the generative image. The accompanying video text oftentimes cannot prescribe enough interactions, fine dynamics, and causality, because of the lack of high-quality data with precise descriptive labels and captions. 
Third and most importantly, \ac{t2v} is more ``fragile'' than \ac{t2i} in terms of end-user perception. In \ac{t2i}, users seem to be tolerant of novel (but less realistic) static results. 
However, they are more sensitive to flaws or artifacts in \ac{t2v} results \cite{kubricht2017intuitive}, especially when generated videos display ill-posed dynamics. Without cherry-picking, it is challenging to ensure that \ac{t2v}-generated videos follow human perceptions of the physical dynamics of the real world.


Similar to researches in \acf{t2i}, there are also large quantities of high-quality data \cite{ramesh2021zero,radford2021learning,ramesh2022hierarchical} that play key roles in the advanced \acf{t2v} state-of-the-art. 
But in comparison to \ac{t2i} cases, such data are, unfortunately, much less readily available on the internet.
Existing public datasets are either of low quality or of limited number, whereas high-quality video-text sets are generally privately owned by commercial organizations.
This fact necessitates the development of computational methods and tools capable of producing high-quality labeled video-text data that follows human intuition in the (semi-) automatic manner.
Since it is hard, if not impossible, to directly enforce physical rules in the space of the projected image (video), we make the data generation 3D.
Then, advanced 3D simulation techniques could be leveraged.
Additionally, offline rendering makes it possible to further obtain photo-realistic videos captured from multiple viewpoints. 
We believe it would potentially extend \acf{t2v} to \acf{t2-3d}, \acf{t2s} and \acf{t2a} domains.


To this end, we propose an automatic pipeline to generate high-resolution, physically realistic animation with descriptive texts. Our method involves state-of-the-art physical simulation frameworks for producing accurate 3D real-world dynamics. To cover a wider range of physical phenomena, we use: (i) \acf{ipc}\cite{li2020incremental}, a robust solid simulation framework that can accurately resolve the intricate contact dynamics for both rigid bodies and deformable objects with guaranteed intersection-free results; (ii) \acf{mpm}\cite{stomakhin2013material,sulsky1995application}, a multi-physics simulation framework that is capable of simulating versatile solid, fluid, and granular materials and multi-physics procedures. We generate various real-world dynamics, such as \textbf{deformations}, \textbf{fractures}, \textbf{collisions}, \textbf{turbulence}, \etc. With commercial-level rendering tools, we also produce high-resolution multi-view videos. Our contributions can be summarized as follows:
\begin{itemize}
    \item We propose a method to automatically generate high-quality 3D physical-realistic animations along with sentences describing the physical phenomena, including a wide spectrum of commonly seen real-world dynamics.
    \item With state-of-the-art physical simulation methods and rendering tools, we are the first to provide high-quality \acf{t2v} and 3D \acf{t2s} datasets, which will widely benefit \acf{t2i}, \acf{t2v}, \acf{t2-3d}, \acf{t2s}, and \acf{t2a} research.
\end{itemize}

\section{Related Work}
\label{sec:relatedwork}
\paragraph{Text-to-Image and Text-to-Video Generation} 
\cite{reed2016generative} is recognized as the pioneer in \acf{t2i} which extends \acf{gan} \cite{goodfellow2020generative} to multi-modal generation. 
Similarly, \cite{zhang2017stackgan,xu2018attngan} apply \ac{gan} variants and further enhance the quality of the generated images with improved image-text alignments.
Other works, such as DALL-E \cite{ramesh2021zero}, formulate the \ac{t2i} problem as a sequence-to-sequence transfer, and incorporate both Transformer and VQVAE for solutions. 
Some follow-up studies show that the results could be further improved by replacing DALL-E components with other deep modules, such as the CLIP latent space in DALLE2 \cite{ramesh2022hierarchical}.
Moreover, the recent success of \acf{ddm} \cite{ho2020denoising, rombach2022high} also improves the generation quality with cascading up-sampling diffusion decoder. \acf{t2v}, on the other hand, falls much behind \acf{t2i} largely due to the lack of \ac{t2v} datasets. 
To address this issue, most previous works \cite{pan2017create, li2018video} produce relatively low-resolution videos in simplified domains.
Some works also attempt to make use of VAE with attention and extend \ac{gan} to achieve \ac{t2v} generation \cite{mittal2017sync}. 
Latest research \cite{wu2021godiva,hong2022cogvideo,singer2022make,ho2022imagen} extends the \ac{t2i} framework to \ac{t2v} by improving modules in diffusion-based \ac{t2i} framework, adding additional attention modules, and making use of both image-text and video-text data.

\vspace{-6pt}
\paragraph{Text-to-3D and Text-to-Animation Generation} As extensions of \ac{t2i}, DreamFusion \cite{poole2022dreamfusion} and \cite{michel2022text2mesh} synthesize 3D meshes from texts. Moreover, DreamField \cite{jain2022zero} generates radiance field with NeRF. \cite{chen2022text2light} uses texts to control lighting conditions in rendering. Besides, several works use CLIP to enable text-to-3D representations. For example, \cite{khalid2022clipmesh} generates mesh and texture in CLIP space; \cite{wang2022clip} incorporates CLIP with NeRF, enabling simple text-editable 3D object manipulation; \cite{tevet2022motionclip} generates human motion from text. \cite{hong2022avatarclip} further applies text-to-3D generation to Avatar.

\vspace{-6pt}
\paragraph{Vision-Language Datasets} Microsoft COCO\cite{lin2014microsoft}, Google concept caption\cite{sharma2018conceptual,changpinyo2021conceptual}, WIT \cite{srinivasan2021wit}, and VisualGenome \cite{krishna2017visual} \etc are most popular fine-labeled image-based \ac{vl} datasets. 
CLVER\cite{johnson2017clevr} is one of the iconic synthetic \ac{vl} datasets. Besides, there are billions of image-text pairs \cite{ordonez2011im2text,plummer2015flickr30k,schuhmann2022laion} available on the internet. However, high-quality video-text datasets are much less available. 
Existing work includes HowTo100M \cite{miech2019howto100m}, which mainly focuses on instructional descriptions, and WebViD \cite{bain2021frozen}, which contains high-quality daily activity video clips. Additionally, MSRVTT \cite{xu2016msr}, MSVD \cite{chen2011collecting}, DiDeMo \cite{hendricks18emnlp}, and ActivityNet \cite{caba2015activitynet} are commonly used, especially for video-language pre-training. Most of them only contain daily human activity without detailed physical dynamics. 

\begin{figure*}[ht]
    \centering
    \includegraphics[width=\linewidth]{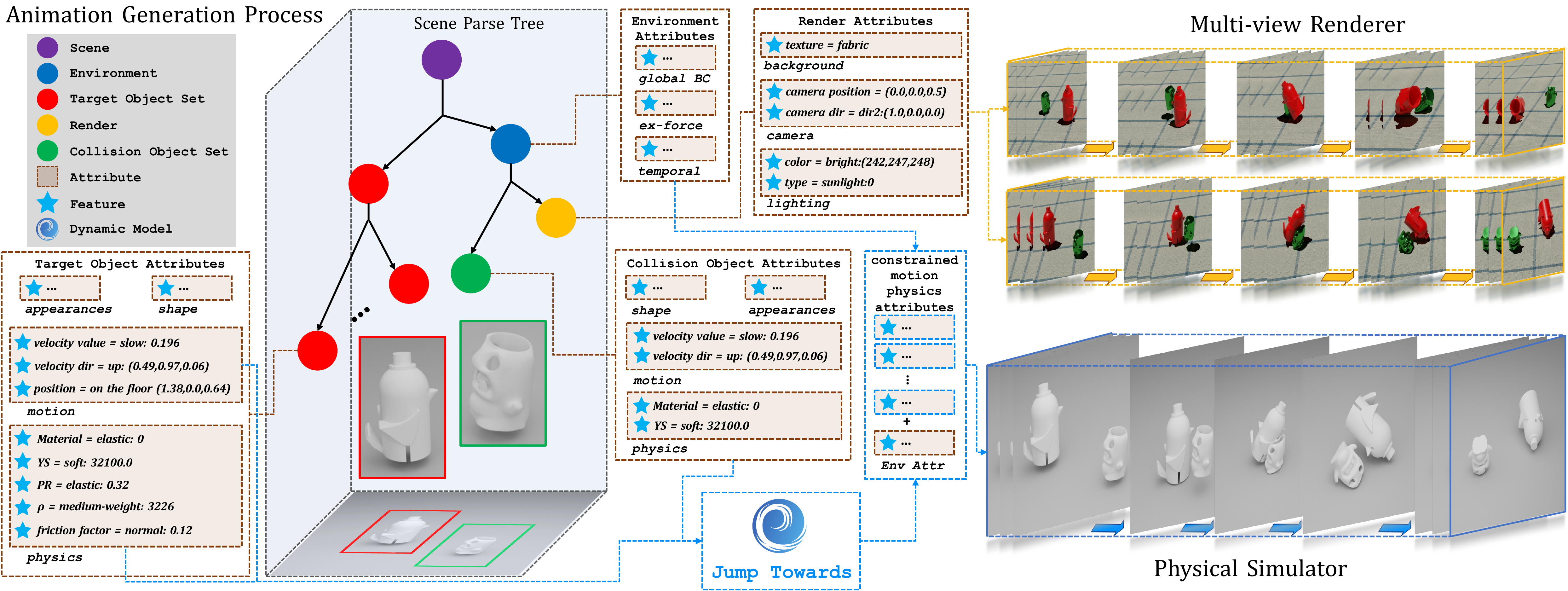}
    \caption{Attributed Scene Grammar define all available setups of the animations. The parse tree of the grammar represents one animation including object-of-interests, dynamics models, simulation environment and renderer. A physical simulation of the sampled animation starts from the pre-processed 3-D object representation, initial motion and physics attributes associated with each object. The attributes of the renderer set up the lightroom, producing multi-view video of the animation.}
    \label{fig:scene}
\end{figure*}
\section{Automatic TPA-Net Generation}
\label{sec:method}
As shown in \autoref{fig:scene}, our work uses a structured representation to construct the physically realistic animation.
In particular, we use attributed stochastic grammar to represent the space of all possible animation setup.
Physics attributes required by physical simulators and renderer are associated as attributes at node-level in the grammatical representation.
Each unique animation setting is a parse tree sampled from the attributed grammar. It contains couples nodes for a set of object-of-interest, a set of collision objects, an environment and a renderer. Different sets of attributes are associated with a node according to the node type. 
The value for each attribute is sampled with certain type of restrictions. Particularly, the sampling process is guided by a randomly selected dynamics model that describes the relationship and mobility of the objects.
First, labels and values of environmental and object-related {\tt attribute}s are sampled based on their intrinsic physical interdependence. Second, relation and motion constraints are applied according to the dynamic model.
After settling all {\tt attribute}s, we dump this data instance to a JSON file and load it to \ac{ipc} or \ac{mpm} simulators, respectively, to simulate elastic or plastic materials.
Later we render the physical simulation results in the renderer to generate photo-realistic videos based on the rendering configurations recorded in the {\tt JSON} file.
Parallel to the simulation process, a stochastic language model is constructed based on the animation representation. A collection of sentences that describe the animation using different descriptors is sampled accordingly.
The subsequent subsections provide additional details.

\subsection{Attributed Scene Grammar}
We use an attributed stochastic grammar as a hierarchical and structured representation to characterize and sample the initial states of a data point in the temporal direction.
Specifically, the stochastic grammar consists of four components: {\tt Target Object Set}, {\tt Collision Object Set}, {\tt Environment} and {\tt Render}. In detail, {\tt Environment} \emph{nodes} represent environmental setups such as boundary conditions, temporal stepping size, external forces, \etc; {\tt Target Object Set} represents the object-of-interest, and {\tt Collision Object Set} represents the collision objects. The material behavior and physical motion of these objects are simulated by the physical simulator; and {\tt Render} \emph{node} contains all the rendering information.
Additionally, each \emph{node} is associated with several {\tt attribute}s, which consist of multiple semantically-related or physically-dependent {\tt feature}s. 
Here, the {\tt feature}, expressed by a qualitative label and a quantitative value, is the atomic semantic model that represents a particular simulation or rendering parameter.
Note that all objects, including collision objects and object-of-interests, have the same {\tt attribute}s to define their shape, mobility, friction, and appearance.
The distinction is that collision objects are treated as boundary conditions with no material type and are configured to remain static (velocity value = zero) over the entire simulation.

\subsection{Dynamic Model}
\label{sec:dynamic}
To capture the characteristics of object motion and the interaction/collisions between specific objects, we provide a dynamic model that determines the type of motion and generates constraints between {\tt feature}s.
Each dynamic model includes a motion type that, semantically speaking, imposes constraints on the velocity value and direction, as well as the initial position of a single or multiple objects. 
Different types of motion require varying numbers of objects, based on user specifications, and may be expanded using prepositions such as ``towards'' to incorporate more objects and collision boundaries.

For instance, the dynamic model {\tt JUMP} depicts a single item with a non-zero velocity and a mostly upward-pointing direction; its initial position should be on top of certain boundaries, such as the floor.
This dynamic model could be extended by adding characteristics such as ``towards'' and ``from'' to introduce a new relationship between the object's motion and other objects or collisions.
Here, the former constrains the velocity direction to point toward additional objects or directions, whereas the latter restricts the initial position to be above the target of ``from''.
Another example is {\tt STRIKE}, which semantically entails the participation of at least two objects with non-zero velocities heading towards some contact center. 
Similarly, the {\tt PUSH} model considers two initially close-by objects, one of which has more kinetic energy and is traveling towards the other.

In practice, dynamic models are manually defined. 
Our dataset generator currently supports several unique models, with {\tt JUMP}, {\tt DROP}, {\tt FLY}, {\tt THROW}, and {\tt SLIDE} describing the motion of at least one object, and {\tt PUSH} and {\tt STRIKE} depicting the action and relationship of several objects.

\subsection{Scenario Sample Process}
To build a concrete scenario, it is necessary to sample 1) the parse tree that represents the simulation components in the domain, 2) the labels and values of {\tt feature}s that characterize the object or environmental properties, and 3) the dynamic model that depicts the motion and relationship of objects and collisions in the simulation domain.
These three stages will be discussed further in depth in the following paragraphs.

\vspace{-6pt}
\paragraph{Sample parse tree.}
At this stage, the number of object-of-interest and collision objects used to sample the stochastic grammar is chosen at random. 
Each sampled object node will contain four {\tt attribute}s that, respectively, describe the object's rendering style, shape, motion, and physical properties.
Then, the components of the simulation domain are decided, and the parse tree will serve as a guide for the dynamic model sampling and language model generation processes.
The sampling results are then output into {\tt JSON} files, with both labels and values recorded, as a general representation of a data point.

\vspace{-6pt}
\paragraph{Sample {\tt feature}s.}
As previously described, each node in the parse tree contains a number of {\tt attribute}s, each of which is comprised of multiple {\tt feature}s that determine a specific simulation or rendering parameter.
In this step, {\tt feature}s are sampled in order to finalize the language and simulation configuration.
We first employ a top-down sampling strategy to collect samples of each feature's label.
Furthermore, label dependencies will be imposed by the semantic relationships between {\tt feature}s managed by identical {\tt attribute}s. 
In the physical property attribute, for instance, we have {\tt feature}s representing the material type, the Young's modulus, and the Poisson ratio separately to depict the object's deformation characteristics. 
The ranges of values for features such as Young's modulus and Poisson ratio are determined by the material type. 
In general, the Young's modulus of elastic bodies is less than that of rigid materials.
Throughout the top-down sampling, all label dependencies are applied.
The feature values are then sampled according to the semantic meaning of the label.

\begin{figure*}[ht]
    \centering
    \includegraphics[width=\linewidth]{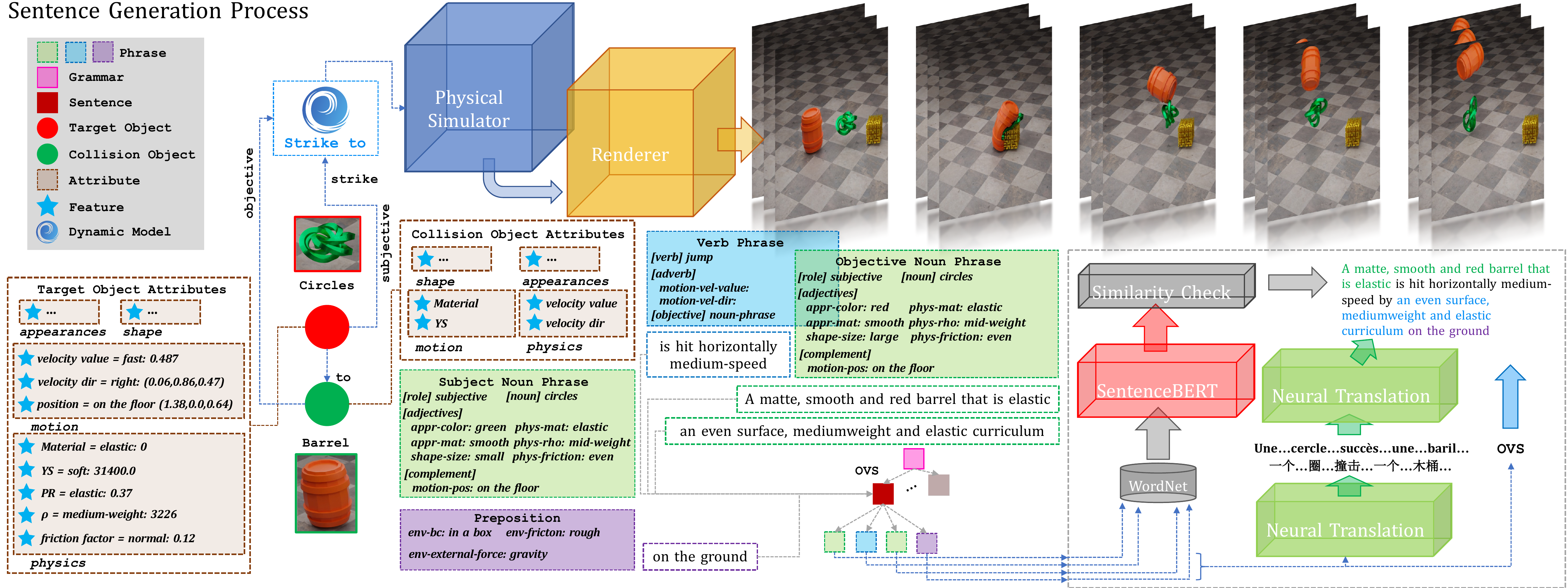}
    \caption{Sentence generation process of animation in TPA-Net. Each object in the animation is associated with a noun phrase. Attributes of the objects determine the adjectives which describe appearance, material, shape \etc. The verb phrase is generated according to the dynamic model. Each pre-defined dynamics has it unique motion and it correspond to a specific text description. A simple grammar is used to compose the sentence. In order to make the generated sentences as diverse as possible, we make use of \ac{nlp} tools, incorporating synonym replacements and neural machine translation.}
    \label{fig:language}
\end{figure*}

After sampling dynamic models (next paragraph), constraints describing the motion and relationship of the object-of-interest and collision objects are sampled.
Consequently, the corresponding {\tt feature}s require updates and resamples.
This time, the constraints are applied to the feature values before the corresponding label is assigned.

\vspace{-6pt}
\paragraph{Sample dynamic models.}
As indicated in \S \ref{sec:dynamic}, each dynamic model has a minimum need for associated objects.
In this step, a dynamic model is picked at random to guarantee that the total number of simulation and collision objects is sufficient to accommodate the motion and relationship portrayed by the current model. 
Then, we randomly choose objects from the scenario parse tree and determine the subjective or objective of the motion. 
Note that the subjective can only be the object-of-interest, while the objective can be any types of objects.
If there are additional free objects or collision items in the scene, we will decide whether to include them as the prepositional characters. 
If this is the case, the current dynamic model will be adjusted to include additional restrictions.

\subsection{Simulation and Rendering}
As previously proposed, once the scene parse tree structure and associated {\tt feature} values have been determined, the data point is captured in {\tt JSON} format.
This output {\tt JSON} file is provided to \acf{ipc}~\cite{li2020incremental} and \acf{mpm}~\cite{qiu2022sparse} simulators according to the material of simulated objects. 
The physical simulators first set the necessary parameters based on the corresponding {\tt feature} values, and then simulate motion and material behaviors such as deformation and fractures until the maximum frame number is reached. 
The output results are then collected by the renderer for high-fidelity rendering using the predefined parameters in the {\tt JSON} file. Our pipeline employs Blender~\cite{blender2018blender} for the automatic rendering process.

Using the intermediate {\tt JSON} description of the scenario, the scene construction and sample, simulation, rendering, and the sentence generation (\S \ref{sec:lan_model} \& \ref{sec:lan_sample} ) phrases can be connected. 
Since the {\tt JSON} file eliminates the data reliance, which is the only dependency among these stages, the methods used at each stage are interchangeable.
As an alternative to our in-house \acf{ipc} and \acf{mpm} simulators, for instance, we can use simulators like NVIDIA's FleX~\cite{macklin2014unified}; similarly, any commercial or non-commercial rendering engine can be utilized for rendering.

\subsection{Language Generation Model}
\label{sec:lan_model}
We construct a tree-structured language model that summarizes the scene parse tree using natural language syntax components.
Here, we investigate the most typical linguistic instance where, in each phrase or sentence, the subjective and objective refer, respectively, to an identical noun word specifying the corresponding scene objects.
As shown in \autoref{fig:language}, sentences are decomposed into multiple sub-sentences, and each sub-sentence consists of three basic components: a noun-phrase for subjective, a verb-phrase for verb and objective, and a global preposition that describes the high-level environmental characteristics.
Noun-phrases delineate a sort of object in the scenario, with a noun and many descriptors pointing to {\tt feature} labels in the parse tree.
In this instance, the noun-phrase object must act as subjective in some dynamic models.
The verb-phrase, on the other hand, consists of a verb indicating the motion of the subjective according to the associated dynamic model and a noun phrase summarizing the objective, if any.

When constructing language model, we examine all of the {\tt feature}s in the parse tree and dynamic model and assign them to language components if they merit being stated in sentences. 
Specifically, the shape {\tt feature} of an object is collected by some noun-phrases as the noun part; the type of dynamic model is mapped to verb in verb-phrases; and the {\tt feature}s linked to the environment node will function as global prepositions. 
Other {\tt feature}s are regarded as adjective descriptors in noun-phrases, with the following exceptions: 
1) Motion {\tt attribute} {\tt feature}s are processed individually. The velocity value and velocity direction are collected as auxiliaries by the verb-phrase, while the position is stored in noun-phrase as a complement.
2) Young's Modulus and Poisson Ratio are neglected since they are too specific to be incorporated into the linguistic model.

\subsection{Generating Random Sentences}
\label{sec:lan_sample}
Along with having a structured language model introduced above, the next step is to generate random sentences to describe the scenario accordingly. 
We divide the problem into several steps which are introduced in the following paragraphs, separately.

\paragraph{Sample Sentence Structure.}
To form the entire sentence, we can finalize the content of each sub-sentence first and connect them with conjunctions such as ``and''.
Here, we consider the case, according to the language model, where sub-sentences consists of subjective, verb, objective and a global preposition.
These components can be connected in multiple ways, i.e., we can choose different component combinations and arrange them in different orders to form the final sentence pattern. For instance, SVOP is the most traditional English sentence structure, while OVS order creates a passive voice (S:subject,V:verb, O:object, P:preposition).
The sentence structure is sampled first when generating concrete clauses.

\vspace{-6pt}
\paragraph{Sample Components.}
With structure defined, we will independently sample each sentence component. 
Generally speaking, it is unnecessary or cumbersome for a natural language sentence to describe every single relevant aspects of a scenario. 
Taking this into account, we select a confined number of descriptors at random to present when producing the final clause. 

In noun-phrases, descriptors might come ahead of the noun as adjectives or after as clauses. 
To choose the candidates for these two portions, we sample two non-overlapping descriptor sets with an arbitrary number of labels from the noun-phrase structure and connect them using commas or conjunctions. 
Note that the complement descriptors can only occur in the clause portion, and the descriptor size can be sampled as zero to indicate that there is no such description portion in the final clause.
The noun-phrase template will be formed by connecting the adjectives-, noun-, and clauses-portion together.
For example, we may sample two adjectives and three clause-descriptors from the subjective noun-phrase in \autoref{fig:language} to create a noun-phrase-template as ``a \adj{blue and matte} \noun{cube} that is \clause{small, elastic and rough}''. 
Here the \adj{blue}, \noun{green} and \clause{brown} words refers to the \adj{adjective-}, \noun{noun-} and \clause{clauses}-portion, respectively.

Similar strategies are employed in the verb-phrase case to handle descriptors. To boost the variety of sentences, we further sample the verb's tense when generating the verb-phrase template. The verb's objective is also sampled as a noun-phrase. As for the preposition portion, a random number of labels coupled by conjunctions are sampled.

\begin{figure}[h]
    \centering
    \includegraphics[width=\linewidth]{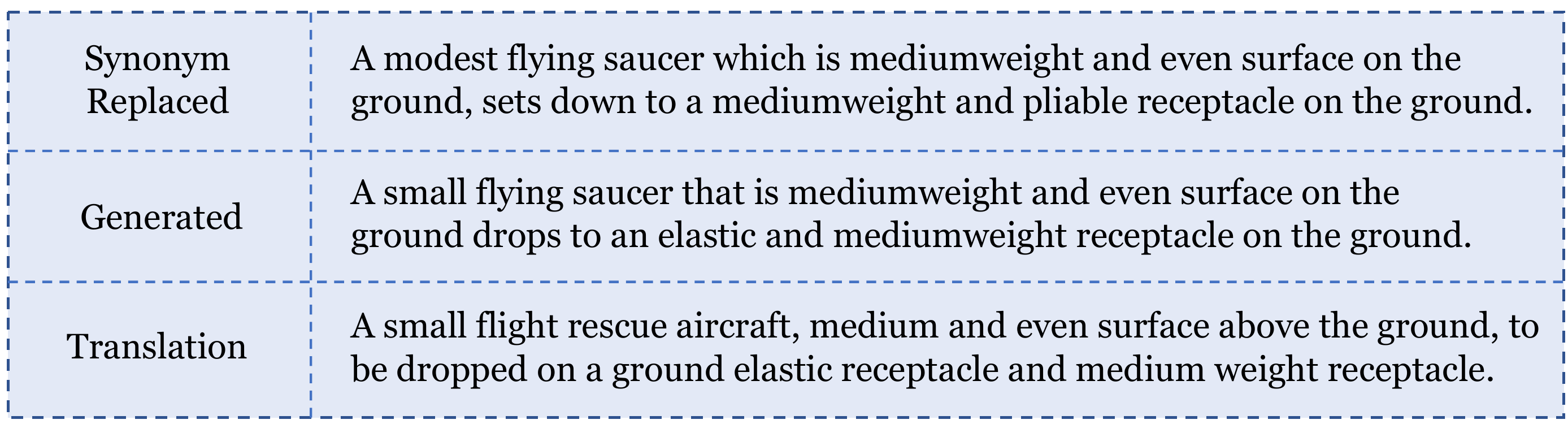}
    \caption{An example of generated sentence, synonym replaced sentence, and translated sentence.}
    \label{fig:sentence}
\end{figure}

\begin{figure*}[ht]
    \centering
    \includegraphics[width=0.81\linewidth]{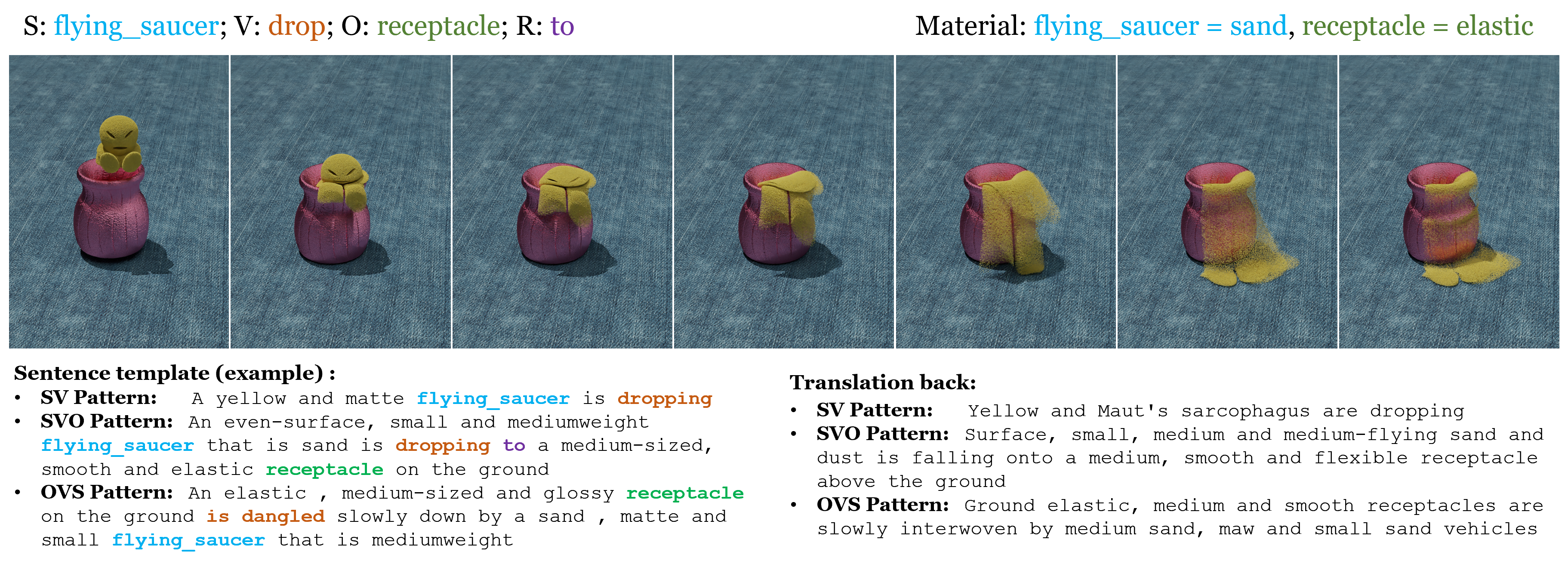}
    \caption{\textbf{Drop dynamics.} A sand \textit{flying\_saucer} (alien) is dropped onto an elastic \textit{receptacle} (something like a vase).}
    \label{fig:drop}
\end{figure*}

\vspace{-6pt}
\paragraph{Sentence Diversity} 
To maximize the variability of the sampled sentences, multiple \acf{nlp} techniques are applied to the clause template.
We begin by substituting \textit{nouns}, \textit{verbs}, \textit{adjectives} and \textit{adverbs} with their corresponding synonyms using WordNet~\cite{miller1998wordnet} and ConceptNet~\cite{speer2017conceptnet}, two of the most widely used lexical knowledge-graphs in \ac{nlp}, to search for synonyms. 
To ensure that the replaced words retain the original semantics, we measure the similarity between the original template sentence and the replacements. 
We encode two sentences using a BERT model and then compute the cosine similarity of the CLS embeddings. 
\cite{reimers2019sentence}.
Multiple levels of semantic analysis are performed to guarantee that the replacement of words does not alter the meaning of the original sentence template.

Second, we deploy techniques for machine translation to further enhance the sentence variety. 
We deploy a neural machine translation (NMT) model based on a transformer to translate the generated sentence into another language and then back to English. 
Incorporating grammars and words from various languages in this manner increases the diversity of the generated sentences.
In practice, we choose French, German, and Chinese as three alternate languages.
\autoref{fig:sentence} shows an example of diverse sentences.

\begin{figure*}[ht]
    \centering
    \includegraphics[width=0.9\linewidth]{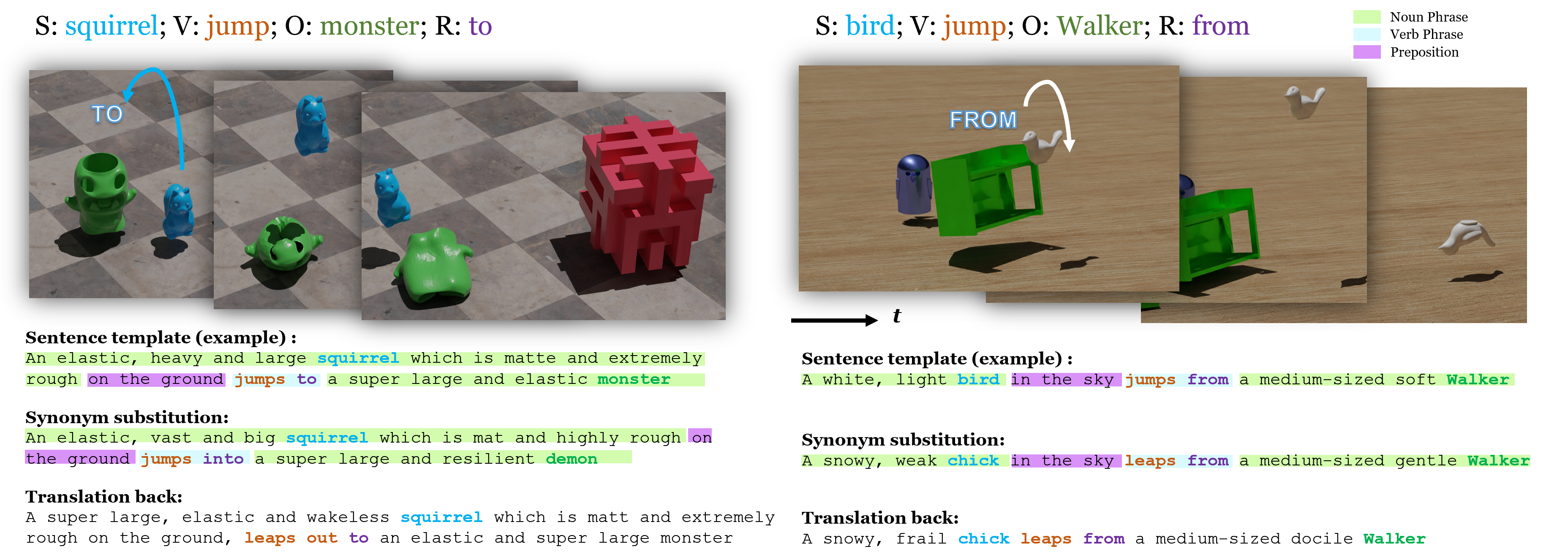}
    \caption{\textbf{Jump dynamics.} The first line depicts the "to" relation between subjective and objective, while the second illustrate "from". Here we show some example of generated sentences. The sentence template is first generated based on {\tt feature} labels; then the words are substituted by synonyms from WordNet and ConceptNet; at last, the sentence are translated to other languages and translated back to get the final descriptive text. Note there can be multiple sentences to describe the same scenario, and not all the descriptors are required to appear in the generated text.}
    \label{fig:jump}
\end{figure*}

\section{Data Processing}
\paragraph{3-D Object Representation} We use Thingi10K\cite{zhou2016thingi10k} as the 3-D object library for the animation generation. 
Pre-processes are needed to fit the input format requirements of \acf{ipc} and \acf{mpm}. Specifically, \ac{ipc} takes a tetrahedral mesh (4 vertices per face) while a 3D volumetric signed distance field is fed into \ac{mpm} simulator.

\vspace{-6pt}
\paragraph{Vocabulary}
In order to get the \textit{nouns} that specify the simulated objects, we pre-process the tags and titles of each data point in Thingi10k to construct the object vocabulary. 
Since the provided labels are unstructured and sometimes customized, we use tools from \acf{nlp} to detect the tags and titles. 
Only those that are \textit{noun}s can be attached to the object vocabulary.
Besides, {\tt feature} labels, are used as the base for constructing the adjective set. 
As for \textit{verbs}, we explicitly define several dynamics, {\tt JUMP}, {\tt DROP}, {\tt FLY}, {\tt THROW}, {\tt SLIDE}, {\tt PUSH} and {\tt STRIKE} (see \S \ref{sec:dynamic}). 
These words forms the basic \textit{verb} vocabulary. 
Except for {\tt PUSH} and {\tt STRIKE}, all \textit{verbs} verbs describe the motion of a single subjective object, with keyword "to" or "from" indicating the relationship between the subjective and the corresponding objective, if any.
{\tt PUSH} and {\tt STRIKE}, on the other hand, intrinsically require at least one objective following the \textit{verb}. In this instance, "to" and "from" will include the third object to indicate additional relationships.
To better illustrate the motions in natural language, we use an adverbial vocabulary and a postposition to enhance the verb with extra information such as the speed of motion, \etc.   

\section{Results}
\label{sec:experiments}
\subsection{Qualitative Results of Text and Animations}
In this section, we show examples and explain the detailed constraints applied for each dynamic case introduced in \S \ref{sec:dynamic}.
In practice, we design eight constraints to depict the relationship among an object's features. \Ie, \textit{less\_eq} and \textit{less} constraints requiring ascending and restrict ascending sequential of involved operands; \textit{larger\_eq} and \textit{larger} in the opposite; \textit{eq} constrain to ensure all operands has the same value; and \textit{same\_dir}, \textit{oppo\_dir} and \textit{similar\_dir} to make the vector operands has the same/opposite/similar direction.
All these constraints take constant values, or vectors, as well as the object and its constrained {\tt feature} as operands, and constraints are applied during the sampling process.

\vspace{-6pt}
\paragraph{Single-object Dynamic.}
As introduced before, we support {\tt JUMP}, {\tt DROP}, {\tt FLY}, {\tt THROW}, {\tt SLIDE} for single-object dynamics. Each of them imposes a different set of constraints.
For example, {\tt Jump} dynamic describes that the subjective object contains a velocity component pointing upwards in the beginning, thus invoking two constraints in our model, \ie, 1) similar velocity direction to the \textit{up} vector, and 2) non-zero velocity value.
{\tt Jump} with a single subjective is sufficient enough to generate sentences with SV sentence structure.

In addition to that, we use "to" and "from" to involve an objective.
As shown in \autoref{fig:jump}, "to" semantically means the subjective is moving towards the objective.
We have two ways to depict this relationship in constraints: 1) the objective is placed at the position where, roughly, the subjective is moving towards; and 2) the subjective should have velocity components pointing to the objective's position.
In practice, we randomly choose one constraint to satisfy in the sampling process.
Similarly, "from" relation constrains the initial position of either the subjective or the objective, to ensure they are placed closely without overlap.

Another example is {\tt DROP}, where it requires the subjective to be placed in the sky, and contains little or no initial velocity value.
If there are any objectives, their initial position should be constrained by the position of the subjective object.
An example is shown in \autoref{fig:drop}.

Similarly, {\tt FLY} and {\tt THROW} impose constraints on the subjective to have non-zero horizontal velocities, while {\tt FLY} further requires the subjective to have positive $y-$ coordinates in the initial position.
{\tt SLIDE}, on the other hand, requires subjective to be placed on top of the ground or attached to some objective objects if there are any.

\vspace{-6pt}
\paragraph{Multi-object Dynamic.}
{\tt PUSH} and {\tt STRIKE}, as multi-object dynamics, requires at least one subjective and one objective object, where the verb, or say the action, will be placed from the subjective to the objective. In this case, "to" and "from" introduce further objects to indicate the velocity direction and initial position settings of the subjective and the objective. More implementation details and corresponding examples are in the supplementary.

\subsection{Qualitative Results of T2I Generation}
In this section, we show some qualitative results of the available \ac{t2i} model in our setting.
We would like to emphasize that there are no open-accessible \acf{t2v} models for us to evaluate the generated data, therefore, we choose to use the open-source \acf{t2i} model, DALLE-mini, to generate images given the texts. One important thing is that we are not focusing on where the model can perfectly generate the image but where it can produce physically realistic images. 

\begin{figure}
    \centering
    \includegraphics[width=0.95\linewidth]{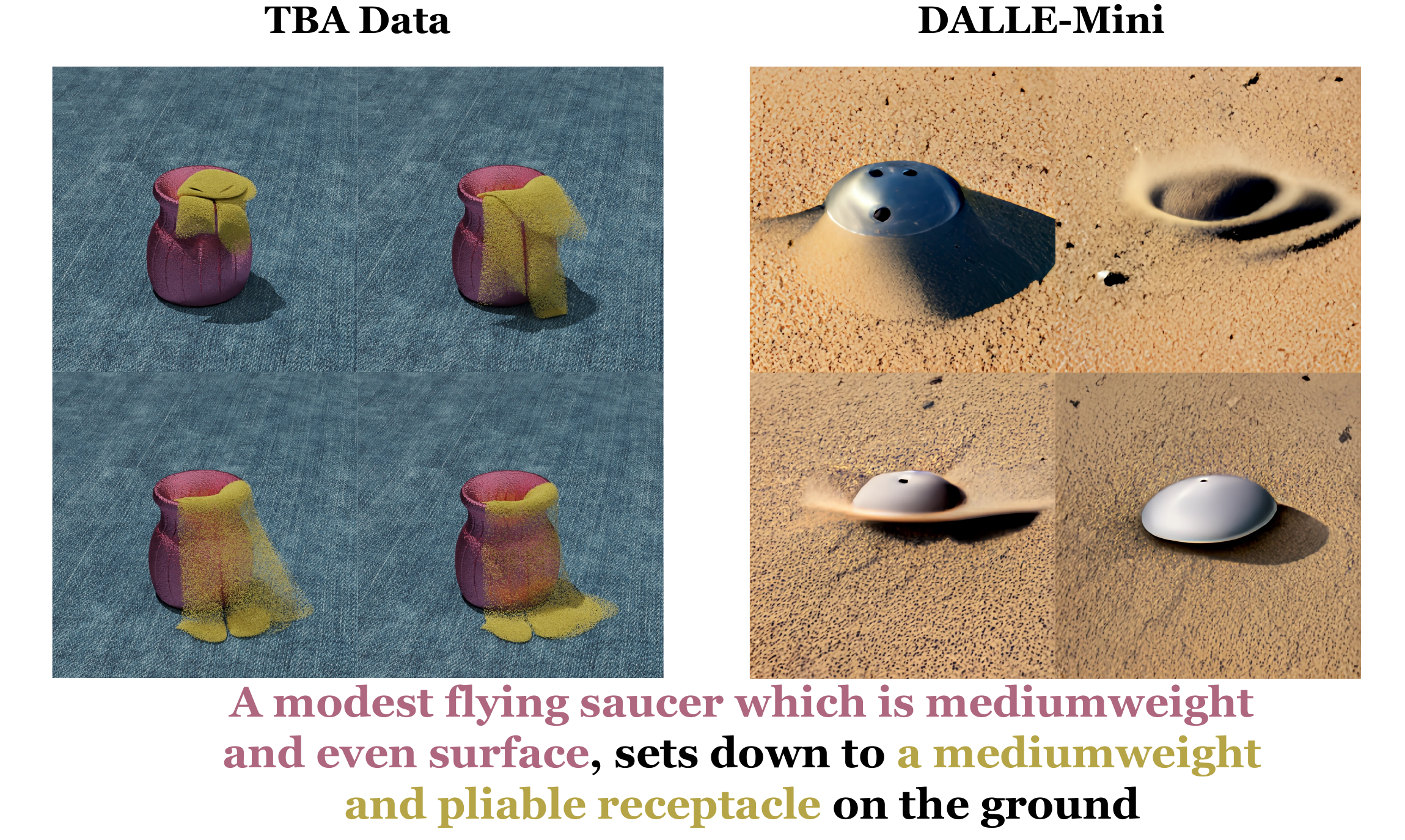}
    \caption{Side-by-side comparison of TBA data and mini-DALLE generated images.}
    \label{fig:dalle1}
\end{figure}

As shown in \autoref{fig:dalle1}, although DALLE-mini can generate images according to the sentence, it is far from physical realism. As mentioned in the \S \ref{sec:intro}, since the all \ac{vl} data do not focus on fine physical details, it is hard for models to capture such subtle but perceivable differences.

\section{Conclusion}
\label{sec:conclusion}
We propose a method to automatically generate text for the physical-based animation method. We show the analysis and qualitative results of our data generation method. Qualitative experiments are also carried out to demonstrate the critical importance of proposing a physically realistic dataset to the multi-modal generation research community.

\newpage
{\small
\bibliographystyle{ieee_fullname}
\bibliography{references}
}

\newpage
\appendix

\begin{table*}[tb]
    \centering    \begin{tabular}{c|l}
        \hline
        Label & Production Rules \\
        \hline
        \hline
        {\tt Scene} & {\tt Scene $\rightarrow$ TarObjSet $\oplus$ Env} \\
        \hline
        {\tt Component-0} & {\tt TarObjSet $\rightarrow$ TargetObj$^+$ $\odot$ TarObjSet$^*$} \\
        & {\tt Env $\rightarrow$ ColObjSet $\oplus$ Render}\\
        \hline
         {\tt Component-*} & {\tt ColObjSet $\rightarrow$ CollisionObj$^*$ $\odot$ ColObjSet$^*$} \\
        & {\tt TarObjSet$^*$ $\rightarrow$ TargetObj$^*$ $\odot$ TarObjSet$^*$} \\
        \hline
        
    \end{tabular}
    \caption{\textbf{Production rule of the scenario stochastic grammar.} Here, {\tt TarObjSet} is short for {\tt Target Object Set} which includes a set of simulated object ({\tt TargetObj}) with potential relationships; {\tt ColObjSet} represents a set of non-movable collision objects ({\tt CollisionObj}) serving as boundary conditions; and {\tt Env} is short for {\tt Environment}.
    Moreover, $\oplus$ represents \textit{and} relation, making the child elements mandatory; while $\odot$ refers to \textit{or} relation to connect optional child nodes; $^+$ means one or more and $^*$ means zero or more.
    }
    \label{tab:scene}
\end{table*}

\section{Examples}
Please visit the following site for live examples:

\url{https://sites.google.com/view/tpa-net} 

\section{Scene Grammar Productions}
We use an attributed stochastic grammar as a hierarchical and structured representation that determines the scenario's content with initial physical parameters and appearance settings.
The grammar is decomposed into multiple levels of components which are sampled according to the production rules defined in \autoref{tab:scene}.
The tree structure itself describes the scenario's content, while the related {\tt attribute}s, which contain numerous {\tt feature}s, specialize the content's characteristics.
\autoref{tab:attrib} presents a list of the {\tt attribute}s and {\tt feature}s designed for each \textit{node}.

\begin{table*}[tb]
    \centering
    \begin{tabular}{c|c|l|l}
    \hline
         \textit{Node} & {\tt Attribute} & {\tt Feature} & {\tt Label Candidates} \\
         \hline
         \hline
         \multirow{6}{*}{Environment} 
             &\multirow{2}{*}{Boundary}  
                   & \textsf{Boundary} & Box, Floor  \\ \cline{3-4}
             & \multirow{2}{*}{Condition}
                   & \textsf{Type}     & Sticky, Slip \\ \cline{3-4} 
             &     & \textsf{Friction Factor} & Smooth, Even Surface, Rough, Extremely rough\\
             \cline{2-4}
             & \multirow{1}{*}{External} 
                & \textsf{Force type} & Gravity, Wind\\ \cline {3-4}
             & \multirow{1}{*}{Force} 
                & \textsl{Force value} & \textit{Dependent on \textsf{Force type}}\\ 
             \cline{2-4}
             & Temporal  
                & \textsf{Total Frame} & Short, Medium, Long \\ \cline{3-4}
         \hline
         \multirow{12}{*}{Object}
            & \multirow{2}{*}{Appearance} 
                & \textsf{Color} & White, Red, Blue, Green, Lime, Orange, Yellow, Pink, Purple ...\\ \cline{3-4}
            &   & \textsf{Material} & Glossy, Matte\\ 
            \cline{2-4}
            & \multirow{2}{*}{Shape} 
                & \textsf{Shape} & Cube, Sphere, Cylinder, Mesh\\ \cline{3-4}
            &   & \textsf{Size} & Small, Medium-sized, Large, Super large \\
            \cline{2-4}
            & \multirow{3}{*}{Motion}  
                & \textsf{Velocity value} & Slow, Medium-speed, Fast \\ \cline{3-4}
            &   & \textsf{Velocity direction} & Up, Down, Right, Left, Forward, Backward, Horizontal, Vertical \\ \cline{3-4}
            &   & \textsf{Initial position} & On the ground, In the sky\\
            \cline{2-4}
            & \multirow{5}{*}{Physics}  
                & \textsf{Material} & Elastic, Rigid, Fluid, Snow, Mud, Sand, Granular\\ \cline{3-4}
            &   & \textsl{Young's Modulus} & \textit{Dependent on \textsf{Material}}; Soft, Moderate-hardness, Hard, Rigid \\ \cline{3-4} 
            &   & \textsl{Poisson Ratio} & \textit{Dependent on \textsf{Material}} Elastic, Rigid\\ \cline{3-4}
            &   & \textsf{density} & Light, Medium-weight, Heavy \\ \cline{3-4}
            &   & \textsf{Friction factor} & Smooth, Even Surface, Rough, Extremely rough \\ 
        \hline
        \multirow{4}{*}{Render} 
            & \multirow{2}{*}{Background} 
                & \textsf{Light} & Bright, Dark \\ \cline{3-4}
            &   & \textsf{Texture} & Preset texture list \\ 
            \cline{2-4}
            & \multirow{2}{*}{Camera}
                & \textsf{Viewpoint} & Preset camera viewpoints \\ \cline{3-4}
            &   & \textsl{Camera} & \textit{Dependent on \textsf{Viewpoint}} \\
            
        \hline
    \end{tabular}
    \caption{\textbf{{\tt Attributes} with {\tt features} associated for each scene \textit{node}.} In the table, independent features are highlighted with the \textsf{sens serif} font, whereas dependent features are labeled with \textsl{italic}. The final column lists examples of candidate labels for each feature. Each label is mapped to a specific value or range of values based on its semantic.}
    \label{tab:attrib}
\end{table*}
\section{Dynamic Model and Constraints}

\subsection{Constraints}
In practice, we design the eight constraints listed below to reveal the relationship among {\tt feature}s of objects.
Every constraint consists of a list of operands "$[o_0, o_1, ..., o_N]$" for the involved {\tt feature}s and/or constants and a list of ID number s"$[n_0, .., n_M]$" to represent the unalterable criteria operand.
Specifically, operand $o_i$ refers a constant value or vector, or a \textit{node}-{\tt attribute}-{\tt feature} pair, in which case the value of the corresponding feature is fetched for computation.

\begin{itemize}
    \item \textit{less\_eq}$([o_0, ..., o_N], [n_0, ..., n_M])$: $o_0 \leq ... \leq o_N$, with $o_{n_0}, ... o_{n_M}$ stays unchanged during resampling.
    \item \textit{less}$([o_0, ..., o_N], [n_0, ..., n_M])$: $o_0 < ...< o_N$, with $o_{n_0}, ... o_{n_M}$ stays unchanged during resampling.
    \item \textit{larger\_eq}$([o_0, ..., o_N], [n_0, ..., n_M])$: $o_0 \geq ... \geq o_N$, with $o_{n_0}, ... o_{n_M}$ stays unchanged during resampling.
    \item \textit{larger}$([o_0, ..., o_N], [n_0, ..., n_M])$: $o_0 > ... > o_N$, with $o_{n_0}, ... o_{n_M}$ stays unchanged during resampling.
    \item \textit{eq}$([o_0, ..., o_N], [n_0, ..., n_M])$: $o_0 = ... = o_N$, with $o_{n_0}, ... o_{n_M}$ stays unchanged during resampling.
    \item \textit{same\_dir}$([o_0, ..., o_N], [n_0])$: $o_i$ must be vectors, and for $\forall i \in [0,...,N]$ the angle between $o_i$ and $o_{n_0}$ is zero. Note that only one criteria operand is present in this constraint.
    \item \textit{oppo\_dir}$([o_0, ..., o_N], [n_0])$: $o_i$ must be vectors, and for $\forall i \in [0,...,N]$ the angle between $o_i$ and $o_{n_0}$ is $180^{\circ}$. Note that only one criteria operand is present in this constraint.
    \item \textit{similar\_dir}$([o_0, ..., o_N], [n_0, ..., n_M], \theta)$: $o_i$ must be vectors, and for $\forall i \in [0,...,N]$, $\forall j \in [0,...,M]$ the angle between $o_i$ and $o_j$ is less or equal to $\theta$. Here, $o_{n_0}, ... o_{n_M}$ stays unchanged during resampling.
\end{itemize}

The constrains are validated after a random sample of features. The non-criteria operands that violate the constraints will be resampled to guarantee the correctness of the relation. If the criteria operands themselves violate the constraint, the resample process will be terminated and errors will be reported.

\subsection{Dynamic Model}
As introduced in the main paper, we have the following dynamic models: {\tt JUMP}, {\tt DROP}, {\tt FLY}, {\tt THROW}, {\tt SLIDE}, {\tt PUSH} and {\tt STRIKE}. 
We summarize the basic constraints required for each single-object dynamic model in \autoref{tab:single_dy} and multi-object dynamics in \autoref{tab:multi_dy}. 
We use "sub" to denote the subjective object on which the dynamic model focuses.
In addition, as shown in the last two columns of the table, objective objects (represented by "obj") are introduced with "from" and "to" relations.
\begin{table*}[]
    \centering
    \begin{tabular}{c|c|l}
    \hline
         Dynamic Model &  Type  & Constraint \\
         \hline
         \hline
         \multirow{6}{*}{{\tt JUMP}} 
            & \multirow{2}{*}{Basic} & 
                \textit{similar\_dir}$([[0,1,0], ($sub, Motion, \textsf{Velocity direction}$)], [0], \theta_0)$\\ \cline{3-3}
            & & \textit{less\_eq}$([v_{min}, ($sub, Motion, \textsf{Velocity value}$)], [0])$ ($v_{min}$ defined by user)\\ \cline{2-3}
            & \multirow{2}{*}{"from"} &
                \textit{eq}$([\mathbf{p}^{\text{gt}}, ($sub, Motion, \textsf{Initial position}$)], [0])$, with $\mathbf{p}^{\text{gt}} = [p^{\text{gt}}_0, p^{\text{gt}}_1, p^{\text{gt}}_2]$ \\
            & & \;\; Here, $p^{\text{gt}}_i = p^{\text{obj}}_i \pm (s^{\text{sub}}_i + s^{\text{obj}}_i + C)*0.5$ for $i \in [0,2]$\\ \cline{2-3}
            & \multirow{2}{*}{"to"} &  
                \textit{similar\_dir}$([\mathbf{d}^{\text{gt}}, ($sub, Motion, \textsf{Velocity direction}$)], [0], \theta_1)$ \\
            & & \;\; Here, $\mathbf{d}^{\text{gt}} = (\mathbf{p}^{\text{obj}} - \mathbf{p}^{\text{sub}}) + \alpha\cdot[0,1,0]$ ($\alpha$ defined by user) \\
         \hline
         
         \multirow{8}{*}{{\tt DROP}} 
            & \multirow{3}{*}{Basic} & 
                \textit{similar\_dir}$([[0,-1,0], ($sub, Motion, \textsf{Velocity direction}$)], [0], \theta_0)$\\ \cline{3-3}
            & & \textit{larger\_eq}$([v_{\text{small}}, ($sub, Motion, \textsf{Velocity value}$)], [0])$ ($v_{\text{small}}$ defined by user)\\ \cline{3-3}
            & & \textit{less\_eq}$([\mathbf{p}^{\text{gt}}, ($sub, Motion, \textsf{Initial position}$)], [0])$, with $\mathbf{p}^{\text{gt}} = [p^{\text{gt}}_0, p^{\text{gt}}_1, p^{\text{gt}}_2]$ \\
            & & \;\; $p^{\text{gt}}_1$ is user-defined threshold; $p^{\text{gt}}_0$ and $p^{\text{gt}}_2$ are the global minimum position \\ \cline{2-3}
            & \multirow{2}{*}{"from"} &
                \textit{eq}$([\mathbf{p}^{\text{gt}}, ($obj, Motion, \textsf{Initial position}$)], [0])$ \\
            & & \;\; Here, $p^{\text{gt}}_i = p^{\text{sub}}_i \pm (s^{\text{sub}}_i + s^{\text{obj}}_i + C)*0.5$ for $i\in[0,2]$ \\ \cline{2-3}
             & \multirow{2}{*}{"to"} &
                \textit{eq}$([\mathbf{p}^{\text{gt}}, ($obj, Motion, \textsf{Initial position}$)], [0])$ \\
            & & \;\; Here, $p^{\text{gt}}_i = p^{\text{sub}}_i \pm (s^{\text{sub}}_i + s^{\text{obj}}_i + C)*0.5$ for $i=0,2$; $p^{\text{gt}}_1 =s^{\text{obj}}_1+C$\\
         \hline
         \multirow{8}{*}{{\tt FLY}} 
            & \multirow{3}{*}{Basic} & 
                \textit{similar\_dir}$([\mathbf{v}_{dir}^{\text{gt}}, ($sub, Motion, \textsf{Velocity direction}$)], [0], \theta_0)$, $\mathbf{v}_{dir}^{\text{gt}}=[C_0, 0, C_1]$ \\ \cline{3-3}
            & & \textit{less\_eq}$([v_{\text{large}}, ($sub, Motion, \textsf{Velocity value}$)], [0])$ ($v_{\text{large}}$ defined by user)\\ \cline{3-3}
            & & \textit{less\_eq}$([\mathbf{p}^{\text{gt}}, ($sub, Motion, \textsf{Initial position}$)], [0])$, with $\mathbf{p}^{\text{gt}} = [p^{\text{gt}}_0, p^{\text{gt}}_1, p^{\text{gt}}_2]$ \\
            & & \;\; $p^{\text{gt}}_1$ is user-defined threshold, $p^{\text{gt}}_0$ and $p^{\text{gt}}_2$ are the global minimum position \\ \cline{2-3}
            & \multirow{2}{*}{"from"} &
                \textit{eq}$([\mathbf{p}^{\text{gt}}, ($obj, Motion, \textsf{Initial position}$)], [0])$ \\
            & & \;\; Here, $p^{\text{gt}}_i = p^{\text{sub}}_i \pm (s^{\text{sub}}_i + s^{\text{obj}}_i + C)*0.5$ for $i\in[0,2]$\\ \cline{2-3}
             & \multirow{2}{*}{"to"} &
                \textit{eq}$([\mathbf{p}^{\text{gt}}, ($obj, Motion, \textsf{Initial position}$)], [0])$ \\
            & & \;\; Here, $\mathbf{p}^{\text{gt}} = \mathbf{p}^{\text{sub}} + \mathbf{s}^{\text{sub}} + v^{\text{sub}}_{value} \cdot \mathbf{v}^{\text{sub}}_{dir} \cdot C$\\
        \hline
         \multirow{8}{*}{{\tt THROW}} 
            & \multirow{1}{*}{Basic} & \multirow{2}{*}{\textit{similar\_dir}$([\mathbf{v}_{dir}^{\text{gt}}, ($sub, Motion, \textsf{Velocity direction}$)], [0], \theta_0)$, $\mathbf{v}_{dir}^{\text{gt}}=[C_0, 1, C_1]$}\\
            & \multirow{1}{*}{(Throw UP)}& \\ \cline{2-3}
            & \multirow{2}{*}{Basic} & 
                \textit{similar\_dir}$([\mathbf{v}_{dir}^{\text{gt}}, ($sub, Motion, \textsf{Velocity direction}$)], [0], \theta_0)$, $\mathbf{v}_{dir}^{\text{gt}}=[C_0, -1, C_1]$\\  \cline{3-3}
            & \multirow{2}{*}{(Throw DOWN)} & 
            \textit{eq}$([\mathbf{p}^{\text{gt}}, ($obj, Motion, \textsf{Initial position}$)], [0])$. \\
            & & \;\; Here, $p^{\text{gt}}_i = p^{\text{sub}}_i \pm (s^{\text{sub}}_i + s^{\text{obj}}_i + C)*0.5$ \\ \cline{2-3}
            & \multirow{2}{*}{"from"} &
                \textit{eq}$([\mathbf{p}^{\text{gt}}, ($obj, Motion, \textsf{Initial position}$)], [0])$ \\
            & & \;\; Here, $p^{\text{gt}}_i = p^{\text{sub}}_i \pm (s^{\text{sub}}_i + s^{\text{obj}}_i + C)*0.5$ for $i\in[0,2]$ \\ \cline{2-3}
             & \multirow{2}{*}{"to"} &
                \textit{eq}$([\mathbf{p}^{\text{gt}}, ($obj, Motion, \textsf{Initial position}$)], [0])$ \\
            & & \;\; Here, $\mathbf{p}^{\text{gt}} = \mathbf{p}^{\text{sub}} + \mathbf{s}^{\text{sub}} + v^{\text{sub}}_{value} \cdot \mathbf{v}^{\text{sub}}_{dir} \cdot C$\\
        \hline
        \multirow{8}{*}{{\tt SLIDE}} 
            & \multirow{1}{*}{Basic} & 
                \textit{similar\_dir}$([\mathbf{v}_{dir}^{\text{gt}}, ($sub, Motion, \textsf{Velocity direction}$)], [0], \theta_0)$, $\mathbf{v}_{dir}^{\text{gt}}=[C_0, 0, C_1]$\\ \cline{2-3}
            & \multirow{2}{*}{Basic} & 
                 \textit{less\_eq}$([\mathbf{p}^{\text{gt-min}}, ($sub, Motion, \textsf{Initial position}$), \mathbf{p}^{\text{gt-max}}], [0, 2])$\\
            &  \multirow{2}{*}{(Not "from")} & \;\; Here $\mathbf{p}^{\text{gt-*}} = [p^{\text{gt-*}}_0, p^{\text{gt-*}}_1, p^{\text{gt-*}}_2]$, $p^{\text{gt-min}}_1 = s^{\text{sub}}_1$, $p^{\text{gt-max}}_1 = s^{\text{sub}}_1+C_{\text{small}}$ \\ 
            & & \;\; $p^{\text{gt-*}}_i$ refers to global * position for $i=0,2$ and * refers to min/max \\ \cline{2-3}
            & \multirow{3}{*}{"from"} &
                \textit{eq}$([\mathbf{p}^{\text{gt}}, ($sub, Motion, \textsf{Initial position}$)], [0])$, with $\mathbf{p}^{\text{gt}} = [p^{\text{gt}}_0, p^{\text{gt}}_1, p^{\text{gt}}_2]$ \\
            & & \;\; Here, $p^{\text{gt}}_1 = p^{\text{obj}}_1 + (s^{\text{sub}}_1 + s^{\text{obj}}_1 + C_{small})*0.5$ \\
            & & \;\; Random sample $p^{\text{gt}}_i \in [p^{\text{obj}}_i - s^{\text{obj}}_i * 0.5, p^{\text{obj}}_i + s^{\text{obj}}_i * 0.5]$ for $i=0,2$\\ \cline{2-3}
             & \multirow{2}{*}{"to"} &
                \textit{eq}$([\mathbf{p}^{\text{gt}}, ($obj, Motion, \textsf{Initial position}$)], [0])$ \\
            & & \;\; Here, $\mathbf{p}^{\text{gt}} = \mathbf{p}^{\text{sub}} + \mathbf{s}^{\text{sub}} + v^{\text{sub}}_{value} \cdot \mathbf{v}^{\text{sub}}_{dir} \cdot C$\\
        \hline
    \end{tabular}
    \caption{\textbf{Constraints in each single-object dynamic model.} In the table, $\mathbf{s}$, $\mathbf{p}$, $v_{value}$ and $\mathbf{v}_{dir}$ refers to object size, initial position, velocity value and velocity direction, separately; $C\geq0, C_0\in[-1,1], C_1\in[-1,1], C_{small}\in[0, 0.1*s^{sub}_{max}]$ represents random noise, and $\pm$ means +/- are chosen randomly in practice.
    All the dynamic model has the basic constraints applied to the objects, with randomly sampled "from", "to", or NONE relation. Note that "from" and "to" relation will be chosen only when there are enough objects in the scenario. 
    Specially, for {\tt THROW} model, we will first sample to decide if it is "Throw UP" or "Throw DOWN" before defining the basic constraints.}
    \label{tab:single_dy}
\end{table*}

\begin{table*}[]
    \centering
    \begin{tabular}{c|c|l}
    \hline
         Dynamic Model &  Type  & Constraint \\
         \hline
         \hline
         \multirow{14}{*}{{\tt PUSH}} 
            & \multirow{4}{*}{Basic} & 
            \textit{similar\_dir}$([\mathbf{v}_{dir}^{\text{gt}}, ($sub, Motion, \textsf{Velocity direction}$)], [0], \theta_0)$ \\ 
            & & \;\; Here $\mathbf{v}_{dir}^{\text{gt}} = \mathbf{p}^{obj} - \mathbf{p}^{\text{gt-sub}}$ \\\cline{3-3}
            & & \textit{less\_eq}$([v_{\text{large}}, ($sub, Motion, \textsf{Velocity value}$)], [0])$ ($v_{\text{large}}$ defined by user)\\ 
                 \cline{3-3}
            & & \textit{larger\_eq}$([v_{\text{small}}, ($obj, Motion, \textsf{Velocity value}$)], [0])$ ($v_{\text{small}}$ defined by user)\\ 
                 \cline{2-3}
            & Basic &     
            \textit{eq}$([\mathbf{p}^{\text{gt-sub}}, ($sub, Motion, \textsf{Initial position}$)], [0])$ \\
            & (Not "from") & \;\; Here, $p^{\text{gt-sub}}_i = p^{\text{obj}}_i \pm (s^{\text{sub}}_i + s^{\text{obj}}_i + C)*0.5$ for $i\in[0,2]$ \\ \cline{2-3}
            & \multirow{6}{*}{"from"} &
                \textit{eq}$([\mathbf{p}^{\text{gt-sub}}, ($sub, Motion, \textsf{Initial position}$)], [0])$ \\
            & & \;\; Here, $p^{\text{gt-sub}}_i = p^{\text{obj-extra}}_i \pm (s^{\text{sub}}_i + s^{\text{obj-extra}}_i + C)*0.5$ for $i\in[0,2]$ \\ \cline{3-3}
            & & \textit{eq}$([\mathbf{p}^{\text{gt-obj}}, ($obj, Motion, \textsf{Initial position}$)], [0])$ \\
            & & \;\; Here, $\mathbf{p}^{\text{gt-obj}} = \mathbf{p}^{\text{gt-sub}}_i + K * \frac{\mathbf{p}^{\text{gt-sub}} - \mathbf{p}^{\text{obj-extra}}}{||\mathbf{p}^{\text{gt-sub}} - \mathbf{p}^{\text{obj-extra}}||}$ \\
            & & \;\; $K = 0.5\cdot\max_{i=0,1,2}(s_i^{\text{sub}}+s_i^{\text{obj}})+C$ \\
            \cline{2-3}
            & \multirow{3}{*}{"to"} &  
                \textit{eq}$([\mathbf{p}^{\text{gt-obj-extra}}, ($obj-extra, Motion, \textsf{Initial position}$)], [0])$ \\
            & & \;\; Here, $\mathbf{p}^{\text{gt-obj-extra}} = \mathbf{p}^{\text{obj}} + K * \mathbf{v}^{\text{sub}}_{dir}$ \\
            & & \;\; $K = 0.5\cdot\max_{i=0,1,2}(s_i^{\text{obj}}+s_i^{\text{obj-extra}})+C$ \\
         \hline
          \multirow{8}{*}{{\tt STRIKE}} 
            & "to" & $\mathbf{p}^{to}=\mathbf{p}^{\text{obj-extra}}$\\ \cline{2-3}
            & Not "to" & Random sample $\mathbf{p}^{to}$\\ \cline{2-3}
            & \multirow{4}{*}{Basic} &     
            \textit{similar\_dir}$([\mathbf{p}^{\text{to}}-\mathbf{p}^{\text{sub}}, ($sub, Motion, \textsf{Velocity direction}$)], [0], \theta_0)$ \\
            & & \textit{similar\_dir}$([\mathbf{p}^{\text{to}}-\mathbf{p}^{\text{obj}}, ($obj, Motion, \textsf{Velocity direction}$)], [0], \theta_0)$ \\
            & & \textit{less\_eq}$([v_{\text{large}}, ($sub, Motion, \textsf{Velocity value}$)], [0])$ ($v_{\text{large}}$ defined by user)\\
            & & \textit{less\_eq}$([v_{\text{large}}, ($obj, Motion, \textsf{Velocity value}$)], [0])$ ($v_{\text{large}}$ defined by user)\\ \cline{2-3}
            & \multirow{2}{*}{"from"} &
                \textit{eq}$([\mathbf{p}^{\text{gt}}, ($obj, Motion, \textsf{Initial position}$)], [0])$ \\
            & & \;\; Here, $p^{\text{gt}}_i = p^{\text{sub}}_i \pm (s^{\text{sub}}_i + s^{\text{obj}}_i + C)*0.5$ for $i\in[0,2]$ \\ \cline{2-3}
         \hline
    \end{tabular}
    \caption{\textbf{Constraints in each multiple-object dynamic model.} These dynamic models are applied to at least two objects, one representing the subjective and the other representing the objective  ("sub" and "obj" in the table). "From" and "to" relation will include another objective object, named "obj-extra" in the table. $C\geq0$ represents random noise, and $\pm$ means +/- are chosen randomly in practice.}
    \label{tab:multi_dy}
\end{table*}

\end{document}